\documentclass[conference]{IEEEtran}
\IEEEoverridecommandlockouts
\usepackage[numbers]{natbib}
\usepackage{amsmath,amssymb,amsfonts}
\usepackage{algorithmic}
\usepackage{graphicx}
\usepackage{textcomp}
\usepackage{url}
\usepackage{xcolor}
\def\BibTeX{{\rm B\kern-.05em{\sc i\kern-.025em b}\kern-.08em
    T\kern-.1667em\lower.7ex\hbox{E}\kern-.125emX}}

\usepackage{linguex}
\usepackage{times}
\usepackage{graphicx,subfig}
\usepackage{tikz}
\usetikzlibrary{decorations}
\usepackage{amsmath}
\usepackage{float}
\usepackage{xspace}

\usepackage{forest}
\usepackage{pgfplots}
\pgfplotsset{compat=1.17}
\usetikzlibrary{trees}
\usetikzlibrary{decorations.pathreplacing}
\usepackage{enumitem}
\usepackage{listings}

\usepackage{hyperref}

\newcommand{\vago}{\texttt{VAGO}\xspace}

\newcommand{\vagon}{\texttt{VAGO-N}\xspace}

\begin{document}

\title{Measuring vagueness and subjectivity in texts: 

from symbolic to neural \vago}

\author{\vspace{0.05in}
\IEEEauthorblockN{Benjamin Icard\IEEEauthorrefmark{1}, Vincent Claveau\IEEEauthorrefmark{2}, Ghislain Atemezing\IEEEauthorrefmark{3} and Paul \'Egré\IEEEauthorrefmark{1}}

\vspace{0.2in}
\IEEEauthorblockA{\IEEEauthorrefmark{1}Institut Jean-Nicod (CNRS, ENS-PSL, EHESS), Paris, France}

\vspace{0.03in}
\IEEEauthorblockA{\IEEEauthorrefmark{2}IRISA-CNRS, Rennes, France \& Direction Générale de l'Armement, Bruz, France}

\vspace{0.03in}
\IEEEauthorblockA{\IEEEauthorrefmark{3}Mondeca, Paris, France}
}

\maketitle

\begin{abstract}
We present a hybrid approach to the automated measurement of vagueness and subjectivity in texts. We first introduce the expert system \vago, we illustrate it on a small benchmark of fact vs. opinion sentences, and then test it on the larger French press corpus FreSaDa to confirm the higher prevalence of subjective markers in satirical vs. regular texts. We then build a neural clone of \vago, based on a BERT-like architecture, trained on the symbolic \vago scores obtained on FreSaDa. Using explainability tools (LIME), we show the interest of this neural version for the enrichment of the lexicons of the symbolic version, and for the production of versions in other languages.
\end{abstract}


\section{Introduction}

 How do we decide whether a statement is factual or whether it reports a mere opinion when we lack access to first-hand knowledge? This question is of central importance in order to assess and enhance information quality on the web and in other media. 
 
 In 2018, PEW Research Center (henceforth PRC) conducted a survey among North-Americans, intended to test the public's ability to recognize statements either as expressing mere opinion or as reporting facts in the news. While each statement in their sample was correctly classified by a majority of participants, only 26\% were able to correctly identify all factual statements as such in the test sample, and only 35\% to categorize all opinion statements as opinion \cite{mitchell2018distinguishing}. Despite asking participants to issue judgments ``regardless of [their] knowledge of the topic", PRC did not give an account of the linguistic cues that a competent language-user ought to rationally identify as conveying objective or subjective information. As argued by Kaiser and Wang in \cite{Kaiser&Wang2021}, however, whether a statement is fact-like or opinion-like depends on ``linguistic packaging".


In order to clarify this issue, 
this paper presents a symbolic tool called \vago, designed to assess informational quality in textual documents, by providing measures of vagueness vs. precision in discourse \cite{van2010not, egre2018vague}, but also of subjectivity vs. objectivity \cite{kennedy2013two}. So far, automatic vagueness detection has been considered in the context of privacy policies, for scoring vagueness at the level of words and sentences \cite{bhatia2016theory}, for predicting whether a word is vague or not based on the vector representation of vague expressions in recurrent neural networks \cite{liu2016modeling}, 
or for generating different degrees of vague sentences using adversarial networks 
\cite{lebanoff2018automatic}, with results further exploited in \cite{lian2023really} to evaluate the support of individual sentences to the vagueness prediction of whole documents. Concerning the detection of opinion, measures of subjectivity have been considered at the level of sentences and documents \cite{alhindi2020fact}. Following \cite{todirascu2019genre}, \cite{escouflaire2022identification} observed that in French articles, some syntactic markers  (e.g. exclamation, interrogation and suspension points) are significantly more prevalent in opinion articles than in factual articles, as for semicolon and colon. Conversely, like \cite{kruger2017classifying} for English newspapers, \cite{escouflaire2022identification} also observed that factual articles in French contained more named entities (e.g. dates, values, percentage) than opinion articles.




Our own approach relies on the observation that a subclass of vague expressions, comprised in particular of multi-dimensional vague adjectives, constitutes a reliable marker of subjectivity \cite{Solt2018, Kaiser&Wang2021}, and therefore of opinion. To leverage it, we combine two distinct methods. On the one hand, we use an expert system called \vago \cite{Guelorget2021combining,icard2022vago,icard&alTALN2023}, which relies on symbolic rules for detecting and measuring lexical vagueness and subjectivity in textual documents. On the other hand, we create a neural counterpart of symbolic \vago, called \vagon, in order to test and enhance the expert system's performance. One of the goals of this method is to extend the results of the expert system to languages other than French and English. But we also present the potential use of \vagon, initially trained for predicting scores through regression, for detecting false information, or fake news.

\section{The Symbolic Tool \vago}

\subsection{Typology of vagueness and measures of detail}

\vago measures vagueness and subjectivity in textual documents based on a lexical database for French and English \cite{datasetvago2022}, with rules for vagueness scoring and vagueness cancellation depending on the syntactic context (see below). Built on a typology derived from \cite{Egre&Icard2018}, this database provides an inventory of vague terms in four categories: approximation vagueness ($V_A$), generality vagueness ($V_G$), degree vagueness ($V_D$), and combinatorial vagueness ($V_C$).

Approximation vagueness primarily concerns modifiers like ``environ" (``approximately''), which relax the truth conditions of the modified expression. Generality vagueness includes determiners like ``certains" (``certain'') and modifiers like ``au plus'' (``at most"). Unlike expressions of approximation, the latter have precise truth conditions. The category of expressions related to degree vagueness and combinatorial vagueness \cite{alston1964philosophy} mainly consists of one-dimensional gradable adjectives (such as ``grand" - ``big", ``vieux" - ``old") and multidimensional gradable adjectives (like ``beau" - ``beautiful", ``intelligent" - ``intelligent", ``bon" - ``good", ``qualifié" - ``qualified"). Expressions of type $V_A$ and $V_G$ are treated as \textit{factual}, while expressions of type $V_D$ and $V_C$ are treated as \textit{subjective} \cite{kennedy2013two,verheyen2018subjectivity,Solt2018}.

In the original version of \vago \cite{icard2022vago}, 
a sentence is considered to be vague if it contains at least one marker of vagueness, and subjective if it contains at least one marker of subjectivity. 
However, the precision of sentences and texts is evaluated only \emph{negatively}: a sentence is precise exactly if it does not contain any vague markers. 
As an example, \vago would assign identical vagueness and subjectivity scores of 1 to the following two sentences, because both contain at least one marker of vagueness/subjectivity (``important'' in (a), ``quickly'' and ``excellent'' in (b)):\footnote{Sentence (a) is an English translation of a sentence taken from the Wikipedia article on Joseph Bonaparte, while sentence (b) is inspired by a false news or ``fake news'' story.}

\ex.[(a)] {\small \textit{\underline{King of Naples} from \underline{1806} to \underline{1808}, then of \underline{Spain} from \underline{1808} to \underline{1813}, he is an \textbf{important} figure in the plan implemented by \underline{Napoleon} to establish the sovereignty of \underline{France} over \underline{continental Europe}.}}

\ex.[(b)] {\small \textit{To \textbf{quickly} cure \underline{Covid-19}, one must take an \textbf{excel\-lent} herbal decoction.}}


Intuitively, however, sentence (a), which contains nine named entities (underlined terms in the sentence), is more informative than sentence (b), which only contains one named entity (``Covid-19''), and therefore provides fewer details than (a). To address this limitation, here the current version of \vago is enriched with a relative \textit{detail score}, based on the proportion of named entities compared to vague expressions.

\subsection{Scores: Vagueness, Subjectivity, Detail}
\label{ssec:ratios}

The current version of \vago is able to measure the scores of vagueness, subjectivity, but also of detail in English and French documents. The detection of vagueness and subjectivity relies on a lexical database, which consisted of 1,640 terms in both languages at the time of the experiments \cite{datasetvago2022}, distributed as follows by vagueness category: $|V_A|$ = 9, $|V_G|$ = 18, $|V_D|$ = 43, and $|V_C|$ = 1570. Regarding the level of detail, the detection is based on identifying named entities (such as people, locations, temporal indications, institutions, and numbers) using the open-source library for Natural Language Processing spaCy.\footnote{\url{https://spacy.io/}} 

For a given sentence $\phi$, its \textit{vagueness score} is defined as the ratio between the number of vague words in $\phi$ and the total number of words in the sentence, written $N_{\phi}$:
\begin{align}
\label{eq: vague}
R_{vagueness}(\phi) =
\frac{\overbrace{|V_D|_{\phi} + |V_C|_{\phi}}^{\text{subjective}} + \overbrace{|V_A|_{\phi} + |V_G|_{\phi}}^{\text{factual}}}{N_{\phi}}
\end{align}

\noindent where $|V_A|{\phi}$, $|V_G|{\phi}$, $|V_D|{\phi}$, and $|V_C|{\phi}$ represent the number of terms in $\phi$ belonging to each of the four vagueness categories (approximation, generality, degree vagueness, and combinatorial vagueness). More specifically, the \textit{subjectivity score} of a sentence $\phi$ is calculated as the ratio between the subjective vague expressions in $\phi$ and the total number of words in $\phi$. Similarly, a factual vagueness score can be calculated with the expressions of generality and approximation (see sections \ref{sec:neuro} and \ref{sec:ext}):
\begin{align}
R_{subjectivity}(\phi) &= \frac{|V_D|_{\phi} + |V_C|_{\phi}}{N_{\phi}}\label{eq: sub} \\ 
R_{factual~vagueness}(\phi) &= \frac{|V_A|_{\phi} + |V_G|_{\phi}}{N_{\phi}}\label{eq: fact}
\end{align}


The \textit{detail score} of a sentence can be defined as the ratio $R_{detail}(\phi)=\frac{|P|_{\phi}}{N_{\phi}}$, where $|P|_{\phi}$ denotes the number of named entities in the sentence (referential terms). By extension, if $|V|_{\phi}$ denotes the number of vague terms in a sentence (across all categories), we define the \textit{detail/vagueness score} of a sentence as the relative proportion of named entities, given by:
\begin{align}
R_{detail/vagueness}(\phi) &=
\frac{|P|_{\phi}}{|P|_{\phi} + |V|_{\phi}}\label{eq:precision}
\end{align}

\smallskip
\noindent In the previous example, we can verify that $R_{detail/vagueness}($a$)=9/10$, while $R_{detail/vagueness}($b$)=1/3$, indicating a higher measure of informativeness for (a).


For a text $T$ (set of sentences), the vagueness scores (including subjective vagueness and factual vagueness) are defined as the proportion of sentences of $T$ whose vagueness score (subjective or factual) is non-zero. The detail/vagueness score of a text $T$ is defined as the average of the $R_{detail/vagueness}$ ratios for each sentence in $T$.

The online version of \vago, available on the Mondeca website,\footnote{\url{https://research.mondeca.com/demo/vago/}. See \cite{icard2022vago,icard&alTALN2023} for the implementation details.} showcases the functionalities of the original version of \vago. The website offers a graphical interface to measure the vagueness and subjectivity scores of texts using two barometers. The first barometer represents the degree of vagueness in a text (defined as $R_{vagueness}(T)$), while the second barometer indicates the extent to which the text expresses an opinion rather than a fact, in other words, the proportion of subjective vocabulary within the text (defined as $R_{subjectivity}(T)$).




\subsection{English \vago on factual versus opinion statements}

In \cite{Guelorget2021combining}, \vago was applied to a large set of more than 28,000 documents from four different corpora. A positive correlation was found between the classification of texts as biased by a CNN based classifier and the vagueness scores computed by \vago. 
In this section, we provide a more analytic perspective by showing the predictions of \vago on the test set used by PRC to evaluate lay people's judgments of fact vs. opinion. Although the set is very limited (10 sentences), it provides a useful benchmark for comparison, in particular because PRC tested the statements on a large sample of participants ($N=5,035$).



Figure \ref{fig:PRCstatements} presents PRC's 10 test sentences, comprised of 5 statements labelled as ``opinion'' and 5 statements labelled as ``factual'' (based on PRC's prescriptive classification). 
The participants recruited by PRC were instructed that a statement should be considered as ``factual" in case it can ``be proved or disproved based on \emph{objective evidence}" (our emphasis), and \emph{regardless of ``whether you think it is accurate or not"}. By contrast, a statement counted as an ``opinion" if ``they thought that it was based on the \emph{values and beliefs} of the journalist or the source making the statement, and could not definitively be proved or disproved based on objective evidence", and \emph{regardless of ``whether you agree with it or not"}\cite{mitchell2018distinguishing} (our emphasis).

\begin{figure}[h]
\begin{small}
\fbox{
\parbox{8.5cm}{
\begin{enumerate}[leftmargin=*]

\item \emph{ISIS lost a \textbf{significant} portion of its territory in Iraq and Syria in 2017.} 
\hfill[\textbf{F}
/\textbf{\textcolor{blue}{O}}]
\smallskip

\item \emph{Immigrants who are in the U.S. illegally have \textbf{some} rights under the Constitution.} 
\hfill[\textbf{F}
/\textbf{F}]
\smallskip

\item \emph{Health care costs per person in the U.S. are the \textbf{highest} in the developed world.}\hfill 
[\textbf{F}
/\textbf{\textcolor{blue}{O}}]
\smallskip

\item \emph{Spending on Social Security, Medicare, and Medicaid make up the \textbf{largest} portion of the U.S. federal budget.}\hfill[\textbf{F}
/\textbf{\textcolor{blue}{O}}]
 \smallskip

\item \emph{{President Barack Obama was born in the United States.}}\hfill
\hfill [\textbf{F}
/\textbf{F}]
\smallskip

\item \emph{Democracy is the \textbf{greatest} form of government.}
\hfill [{\textbf{O}
/\textbf{O}}]
\smallskip

\item \emph{Government is \textbf{almost} \textbf{always} \textbf{wasteful} and \textbf{inefficient}.} 
\hfill [\textbf{O}
/\textbf{{O}}]
\smallskip

\item  \emph{Increasing the federal minimum wage to \$15 an hour is \textbf{essential} for the health of the U.S. economy.}  
\hfill [\textbf{O}
/\textbf{O}]
\smallskip

\item \emph{Immigrants who are in the U.S. illegally are a \textbf{very} \textbf{big} problem for the country today.}  
\hfill [\textbf{O}
/\textbf{O}]
\smallskip

\item \emph{Abortion \textbf{should} be legal in \textbf{most} cases.} 
\hfill[\textbf{O}
/\textbf{O}]

\end{enumerate}}}
\vspace*{-0.15cm}
\end{small}
\caption{PRC's sentences, grouped by category (F=fact; O=opinion). Within brackets: PRC's classification (left) vs. \vago's (right). In blue: cases in which \vago\ differs. Boldfaced expressions: \vago\ entries detected.
}
\label{fig:PRCstatements}
\end{figure}


While the proportion of fact vs. opinion answers varied depending on the sentence, for each statement the majority of participants agreed with the classification made by PRC (mode ranging from 54\% to 77\% for factual statements, and between 68\% and 80\% for opinion statements), giving support to the PRC labels.



\begin{figure}[!htb]
  \centering
  \includegraphics[width=0.44\textwidth]{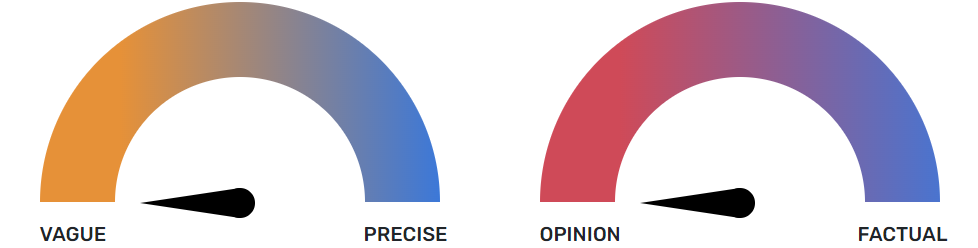}

\medskip
  \includegraphics[width=0.43\textwidth]{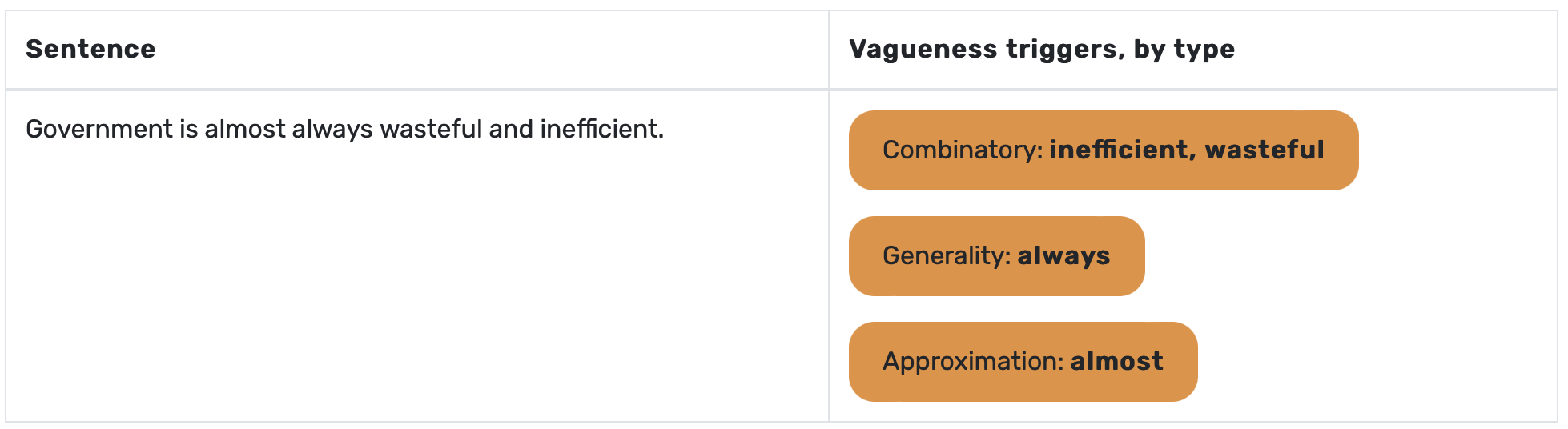}
  \caption{\vago's results for the sentence ``\emph{Government is almost always wasteful and inefficient}". \vago scores are binarized for sentences, but yield intermediate values for larger texts.}
  \label{fig:VAGOresults}
\end{figure}

Because participants were asked to decide whether a statement is factual or opinion regardless of world knowledge and personal beliefs, this justifies looking at the linguistic cues they should be aware of to solve the task.
The category ascribed by \vago to each particular statement is reported in Figure \ref{fig:PRCstatements} (right label). The statements for which \vago's classification differs from the PRC classification are highlighted in blue. In Figure \ref{fig:VAGOresults} are displayed the results of \vago's classification for statement 4, ``\emph{Government is almost always wasteful and inefficient}". Note that for a single sentence, the barometers necessarily take categorical values, intermediate values are only obtained over larger texts.

As shown in Figure \ref{fig:PRCstatements}, \vago classifies eight statements as opinion, and only two as factual. More specifically, the two sentences classified as factual by \vago (2 and 5) are indeed factual according to the PRC's criterion. Conversely, the five sentences labelled ``opinion" by PRC (6-10) are classified as opinions by \vago. This means \vago uses a stricter criterion on what to count as factual, and a laxer one on what to count as opinion relative to this sample. 


According to \vago, sentences 1, 3, 4, 6-10, are all opinion statements since they contain at least one marker of subjectivity (see Figure \ref{fig:PRCstatements}). But, as with the sentences (a) and (b) given above, some of those statements are more informative than others since they contain more named entities. The measure of detail/vagueness we introduced in subsection \ref{ssec:ratios} helps distinguish those sentences in terms of informativity. For instance, although sentences 4 and 9 both receive a score of subjectivity equal to 1, $R_{detail/vagueness}(4) = 4/5$ while $R_{detail/vagueness}(9) = 1/3$.



More specifically, four of the five statements marked as opinion by PRC are identified as such by \vago based on the occurrence of an expression of type $V_C$ (``great", ``wasteful", ``inefficient", ``essential", ``big") as well as $V_D$ (``very"). Our hypothesis is that participants relied on those items to decide that the statements convey subjective values or beliefs.


Looking more specifically at the statements for which the classifications diverge, we can see that in statements 3 and 4, \vago classifies the superlatives ``highest'' and ``largest'' as elements of the category $V_D$. While \vago\ has rules of vagueness cancellation for comparatives in the category $V_D$ (``taller'') and for measure phrases (``5 feet tall''), it does not cancel vagueness in superlatives currently. For ``greatest'' in 6, this is as it should be, since even the superlative leaves room for subjective disagreement, but for ``highest'' and ``largest'', the interpretation seems objective and factual. In the case of 1, ``significant" is the pivotal element behind \vago's classification of the sentence as opinion. We note that 30\% of participants rated the sentence as opinion, possibly relying on the fact that what counts as ``significant" is a matter of interpersonal disagreement. Finally, for sentences 2 and 5 classified as fact by \vago and by PRC alike, 5 is the only sentence categorized as precise by \vago, 2 is vague but ``some'' is $V_G$ and not counted as a marker of subjectivity. 



\subsection{French \vago on regular versus satirical press articles}

\begin{figure}[htb]
\centering
\includegraphics[scale=0.26]{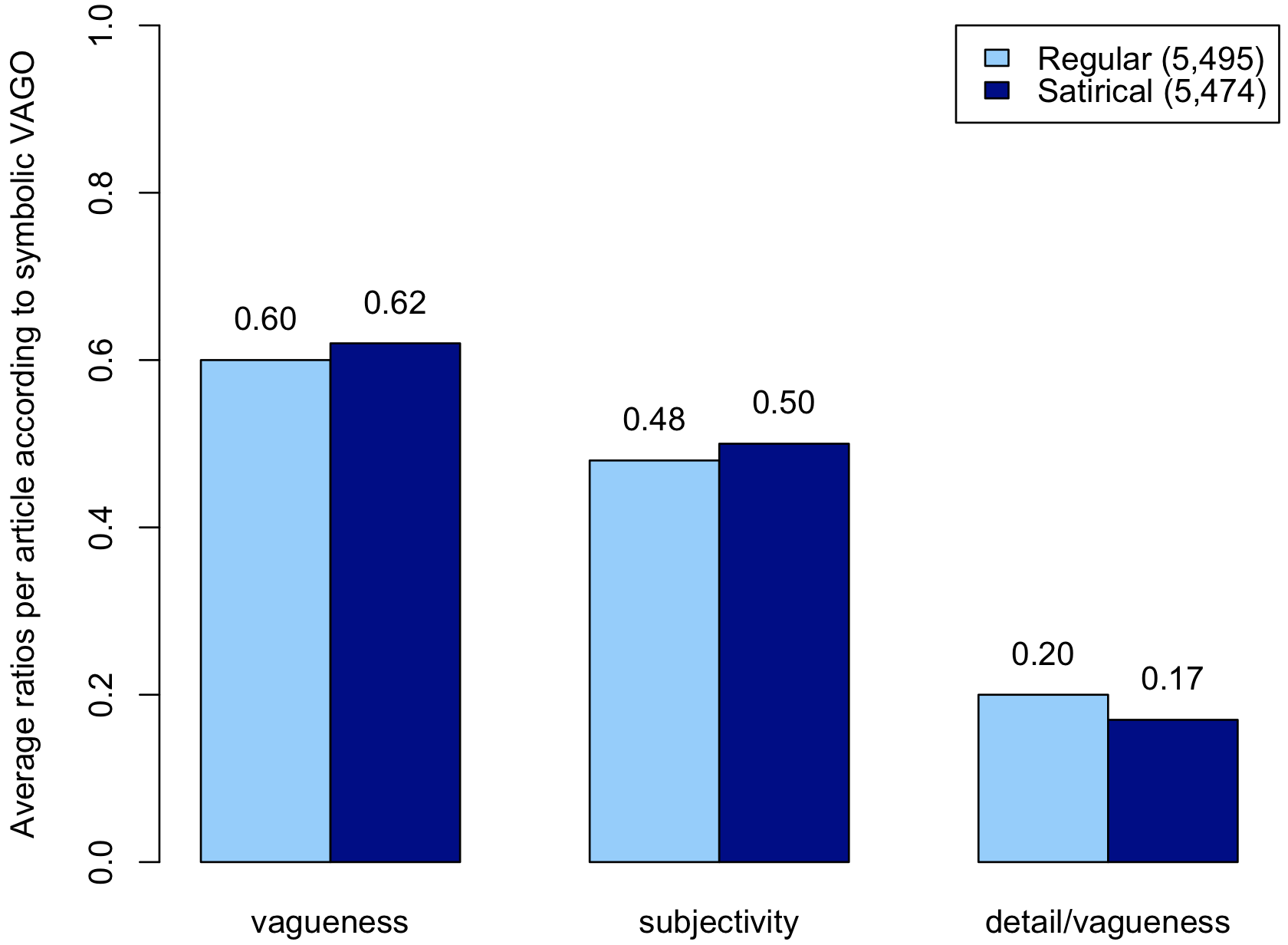}
\caption{Average ratio scores per article of the FreSaDa corpus (French) according to \vago.}
\label{fig:fresada}
\end{figure}

To scale up analytic intuitions, \vago was tested on the French corpus ``FreSaDa''\footnote{\url{https://github.com/adrianchifu/FreSaDa}} \cite{IonescuChifu2021IJCNN}, consisting of 11,570 press articles divided into two supposed homogeneous classes: 5,648 ``regular'' articles from the general French press, not presumed to be false, versus 5,922 ``satirical'' articles explicitly including false or made up content. Within the total corpus, \vago processed 10,969 out of the initial 11,570 articles, with the remaining 601 articles excluded due to inappropriate format for analysis (isolated words, keywords, incomplete sentences, etc.). The results provided by \vago are reported in Figure \ref{fig:fresada}.

According to \vago, the articles in the satirical corpus are significantly \emph{more vague} ($p = 4.99\times10^{-11}$), \emph{more subjective} ($p = 1.69\times10^{-9}$), and \emph{less detailed} ($p = 3.36\times10^{-22}$) than the articles in the regular press corpus (scores calculated per text; two-tailed t-tests, $\alpha = 0.05$, with Bonferroni correction). These results align with expectations and support the findings previously obtained with \vago on English texts \cite{Guelorget2021combining}.

To further exemplify the interest of these ratios, we conducted a simple classification experiment: the goal was to distinguish between the two types of documents in the FreSaDa corpus solely based on the vagueness, subjectivity and detail/vagueness scores. For a given text of the corpus, the ratios of each of its sentences were obtained with \vago; the minimum, median, average and maximum of each ratio on all the sentences were computed and used as input to an XGBoost classifier \cite{XGBoost} (300 rounds, max\_depth=8).
The accuracy obtained with different sizes of the training set 
showed that the ratios scores are effective cues to classify such documents as being either \textit{regular} or \textit{satirical}. Moreover, 
very few examples were necessary to yield high accuracy, the plateau being reached at 500 documents. 



\section{The neural version \vagon}
\label{sec:neuro}


\subsection{Training of the \vago clone}

We built a neural version of the symbolic \vago called \vagon, based on combining a BERT \cite{devlin_bert:_2018} architecture with a regression layer and an MSE loss function to predict a score of vagueness for sentences. In the present experiment, we tested both subjective and factual vagueness, but for the sake of completeness, we also tested the prediction of the $R_{detail/vaguenesss}$ score. However, since this score can be more simply computed from a named entity recognition system, we did not return to it in those experiments. As in a distillation task, \vago was used to associate a vagueness score to each sentence in the FreSaDa corpus and thus to train a neural system. 

In the experiments reported in the following sections, 106,000 sentences out of 141,137 were randomly selected within the 10,969 articles of the FreSaDa corpus processed by \vago, and divided into a training set (85,000 sentences) and a test set (21,000 sentences). We used a RoBERTa Large model (\textit{Batch Size}=30; \textit{Learning Rate}=1e-6; \textit{Epochs}=20); experiments not detailed here with a CamemBERT model \cite{martin2019camembert} provided slightly lower results.

Performance is reported in Table~\ref{tab:resVAGO-N} with standard regression measures: root-mean-square error (RMSE), coefficient of determination ($R^2$), mean absolute error (MAE) and median absolute error (MedAE). All these measures show that \vagon replicates the symbolic \vago scores with high accuracy. The task of detecting subjectivity seems a little more difficult than in the case of factual vagueness detection.

\begin{table}[htb]
\begin{center}
\begin{tabular}[t]{l|c|c|c|c}
                      & RMSE    & $R^2$       & MAE & MedAE  \\
        \hline
     subjective vagueness & 0.022063 & 0.859897 & 0.014518 & 0.009488 \\
     factual vagueness   & 0.008745 & 0.949339 & 0.004124 & 0.001730 \\ 
     detail/vague          & 0.097008 & 0.882543 & 0.051396 & 0.012367 \\
    \end{tabular}
\end{center}
\caption{\label{tab:resVAGO-N} Regression results of \vagon for scores of subjective vagueness, factual vagueness and detail/vagueness concerning the sentences of the FreSaDa corpus (French).}
\end{table}

\subsection{Comparison of the versions of \vago}

\begin{figure}[htb]
\begin{center}
\includegraphics[width=\columnwidth]{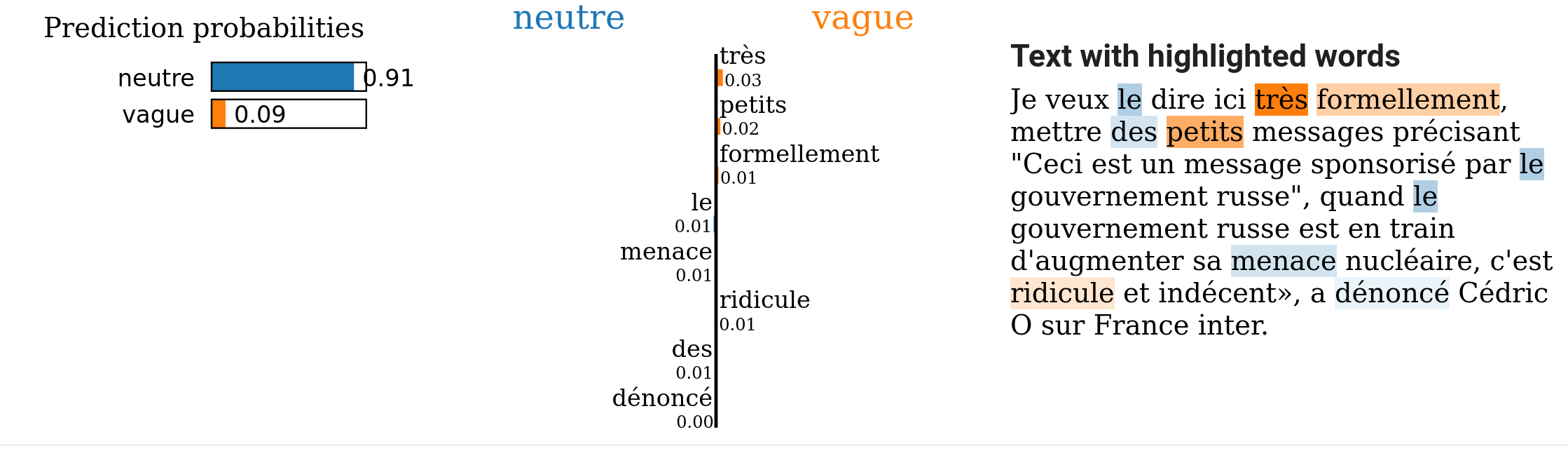}
\caption{\label{fig:LIME1}Example of a LIME output on a French sentence from the FreSaDa corpus processed by \vagon. The category labeled as ``neutre'' corresponds to the inverse category of vagueness, contributing negatively to the vagueness score.}
\end{center}
\end{figure}

The previous quantitative evaluation indicates that \vagon replicates the general behavior of \vago quite faithfully. From a qualitative perspective, we aim to verify here that this neural version relies on the same lexical cues as the symbolic version. For this purpose, we use the explainability tool LIME \cite{LIME}. Applied to the outputs of \vagon, LIME identifies the tokens that contribute the most (or the least) to the vagueness score of a given text. In Figure~\ref{fig:LIME1}, we provide an example of LIME output on a French sentence for subjective vagueness. Using this tool, we examine the cases where the predictions of vagueness scores by \vagon diverge the most from those of \vago.

The study of these error cases puts emphasis on several aspects. Regarding the differences between \vago and \vagon for the prediction of factual vagueness, the vast majority of terms identified as contributing the most to the prediction of \vagon are already present in the French lexicon of \vago, both in case of generality vagueness (e.g., \emph{``tout/tous/toutes'', ``jamais'', ``ou'', ``général'', ``quelques'', ``certains/certaines''}), and in case of approximation vagueness (\emph{``environ'', ``presque''}). Those indicators of factual vagueness are correct, but their weight in the final score of \vagon differs from the calculation performed by the original \vago, which does not put any weight on lexical items. This difference in weight 
may result from words of morpho-syntactic categories not considered by \vago, 
which may amplify or diminish the resulting factual vagueness score, according to the neural network. 


Similarly, in the case of subjective vagueness, LIME applied to \vagon identifies adjectives and terms of excess that were already present in the extant \vago lexicon, either within the category of combinatorial vagueness (e.g., \emph{``négatif'', ``affirmatif'', ``intéressant'', ``fortement'', ``chiant'', ``difficile'', ``probablement'', ``vrai'', ``stupide'', ``vraiment''}), or within the category of degree vagueness (\emph{``petit''}). But LIME also identifies other adjectives carrying combinatorial vagueness that are not yet in the lexicon and should be included (e.g., \emph{``durable'', ``particulièrement'', ``ringard'', ``actuel''}), with some exceptions (``\emph{sabbatique}''). 

\section{Extensions of the \vago Approaches}
\label{sec:ext}


\subsection{Validation and enrichment of the symbolic \vago}
\label{ssec: valid}

For each token $t$ in a text, LIME provides a score for the contribution of $t$ to the vagueness prediction (subjective or factual) in the text, which we denote $c_{occ}(t)$. In the case of a sentence, the higher the $c_{occ}(t)$ of a term $t$ occurring in the sentence, the more positively $t$ contributes to the vagueness score of that sentence.

By applying LIME to the 141,137 sentences of the FreSaDa corpus 
processed by \vago and exploited in \vagon, 
we collected the contribution scores $c_{occ}$ of all occurrences of all the tokens within these sentences. To obtain a global score $c_{tok}(t)$ per token $t$, we summed and normalized the $c_{occ}$ by the total number of occurrences of each token noted here $|occ_t|$: $c_{tok}(t) = \frac{1}{|occ_t|} \sum_{o \in occ_t} c_{occ}(o)$. 
Our hypothesis was that terms from the \vago lexicon should be prioritized among tokens receiving the highest $c_{tok}$. To test it, we calculated the statistical precision (proportion of \vago\ instances retrieved) on the list of tokens ordered by decreasing $c_{tok}$.
Note that a token is taken into account if it is an inflection of a term in the \vago lexicon. The results are listed in Table~\ref{tab:prec-LIME-fr}, where P@$k$ represents the precision over the first $k$ tokens in the list.

\begin{figure}[htb]
\centering
\includegraphics[width=0.33\textwidth]{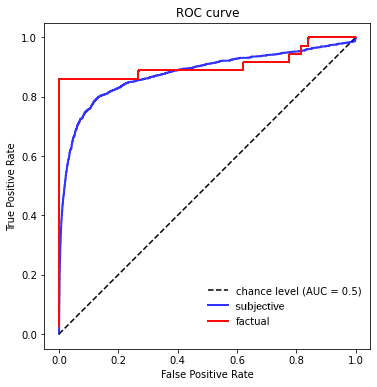}

\caption{\label{fig:ROC-fr} ROC curve of $c_{tok}$ as an indicator of its presence in the French lexicon of \vago.}
\end{figure}

Figure~\ref{fig:ROC-fr} shows the ROC curve relating the $c_{tok}$ score to presence in the French \vago lexicon. These results support our hypothesis. Although \vagon has been only trained on sentences and their scores, it is able to reconstruct the lexicon at the core of the symbolic version.

\begin{table}[htb]
\begin{center}
\begin{tabular}[t]{l|c|c|c|c|c|c}
             vagueness type     & P@5  & P@10  & P@20 & P@30 & P@100 & P@200  \\\hline
    subjective &  1.00 & 1.00  & 0.95 & 0.93 &  0.81 & 0.79 \\
    factual     &  1.00 & 1.00  & 1.00 & 0.93 &  0.31 & 0.16 \\ 
    \end{tabular}
\end{center}
\caption{\label{tab:prec-LIME-fr}Comparison of the precision at different thresholds for the French tokens of the \vago lexicon ordered by $c_{tok}$.}
\end{table}

Besides, we examined the 100 tokens with the highest $c_{tok}$ for subjective vagueness: in addition to the 81 well present in the lexicon, a few verbal forms are listed. Although considered false positives (current \vago lexicons only list adjectives and adverbs), their relevance can be debated. In the remaining tokens, seven absent words were validated as relevant and worthy of inclusion in the lexicons. This list contains four adverbs (\emph{``également'', ``seulement'', ``particulièrement'', ``clairement''}) for which the \vago database contains the root adjectives in two cases (\emph{``particulier'', ``clair''}), an action verb (\emph{``faire''}), an adjective that can also be a noun (\emph{``droit/droite''}), and a noun (\emph{``nombre''}). 
We also noted the detection of non-standard forms of terms present in the lexicon (\emph{``difficille''}, \emph{``pauv''}), illustrating the robustness of the neural approach on noisy text (typo, abbreviation, etc.). The results validate the neural clone \vagon, which retrieves the lexical cues of the expert system \vago, while identifying new or non-standard forms.

\subsection{Developing multilingual versions of \vagon}

Developing symbolic versions of \vago for other languages requires vague lexicons to be available in the target languages. However, automatic translation of these lexicons is not possible due to the idiomatic and out-of-context nature of the lists of expression involved. 

That being said, it is possible to translate the \vagon training set by relying on the following assumption: vagueness scores, in particular subjective vagueness and factual vagueness scores, are preserved from the source language into the target language. To begin with, the FreSaDa corpus was translated from French into English using the Helsinki-NLP/opus-mt-fr-en\footnote{\url{https://huggingface.co/Helsinki-NLP/opus-mt-fr-en}} model  \cite{TiedemannThottingal:EAMT2020}. Next, \vagon was trained to predict subjective vagueness and factual vagueness scores on this English corpus (using the same hyper-parameters as for training \vagon on French). The regression results are similar to those obtained for French, and presented in Table~\ref{tab:resVAGO-N-en}. 

\begin{table}[tb]
\begin{center}
\begin{tabular}[t]{l|c|c|c|c}
      vagueness type & RMSE     & $R^2$       & MAE & MedAE  \\
        \hline
    subjective  & 0.031801 & 0.708915 & 0.022865 & 0.016807 \\
    factual    & 0.016990 & 0.808772 & 0.009582 & 0.004172 \\ 
    \end{tabular}
\end{center}
\caption{\label{tab:resVAGO-N-en} Regression results of \vagon for scores of subjective and factual vagueness concerning the English version of the FreSaDa sentences.}
\end{table}

\begin{figure}[htb]
\centering

\includegraphics[width=0.33\textwidth]{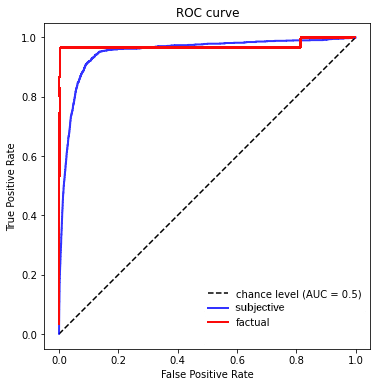}


\caption{\label{fig:ROC-en}ROC curve of $c_{tok}$ as an indicator of its presence in the English lexicon of \vago.}
\end{figure}

Applying the same approach as in subsection \ref{ssec: valid}, we isolated the list of tokens ordered by decreasing $c_{tok}$, then compared it to the English lexicon of \vago, which serves as ground-truth. The accuracy of this list measured at different thresholds is reported in Table~\ref{tab:prec-LIME-en}. The corresponding ROC curves are presented in Figure~\ref{fig:ROC-en}.

\begin{table}[tb]
\begin{center}
\begin{tabular}[t]{l|c|c|c|c|c|c}
               vagueness type        & P@5  & P@10  & P@20 & P@30 & P@100 & P@200  \\
        \hline
     subjective  & 0.80  &  0.90 & 0.90 & 0.93 & 0.90  & 0.84 \\
     factual   &  1.00  &  0.80 & 0.55 & 0.50 & 0.26  & 0.14   \\ 
    \end{tabular}
\end{center}
\caption{\label{tab:prec-LIME-en} Comparison of the precision at different thresholds for the list of English tokens ordered by $c_{tok}$ as a function of the \vago lexicon.}
\end{table}

We also collected the 100 most vagueness-prone English terms according to \vagon. These terms were compared with those in the extant English lexicon of the symbolic version: 90 terms were already included in the \vago English lexicon. Among the highest-ranking terms from the top 100 that do not appear in \vago, we found five adjectives or adverbs that could appear in the combinatorial vagueness category (\emph{``likely'', ``full'', ``complicated'', ``frankly'', ``enough''}), one modal verb (``\emph{must''}), which it also makes sense to include given the presence of \emph{``should''} in the \vago lexicon. Four terms, on the other hand, are not clearly vague (\emph{``course'', ``lost'', ``lose'' and ``finally''}), with the possible exception of ``course'' (occurrences in ``of course'', the use of which is subjective). Of the next 100 terms, all those not included in the \vago lexicon are adjectives that can be included in the $V_C$ category (\emph{``worse'', ``complex''}, etc.). 



\section{Conclusion}

In this article, we first presented \vago, a structured lexicon and rule-based system which calculates scores of vagueness, subjectivity, and detail/vagueness for texts. We then created a neural clone, \vagon, based on BERT. Unlike \vago, \vagon is trained solely on \vago scores, without knowledge of the lexicon underlying this symbolic version. Using LIME, the terms with the highest contribution to \vagon scores turn out to be either terms that already appear in \vago, or terms mostly susceptible to appear in it. This suggests that the decisions of \vagon are largely explained by lexical items identified by the symbolic \vago. Once trained, \vagon can be used to complete the extant lexicons of the symbolic \vago. But it can also be used to produce neural versions of \vago in other languages and to generate lexicons for symbolic versions in these languages.



More work remains to be done. We plan to measure the genericity of our approach by masking the named entities in a text to see whether the \vagon scores remain stable before and after this operation, in order to determine the proportion of the \vago lexicon (which does not contain named entities) actually influencing the decision of \vagon. 

\enlargethispage{.25cm}

On a more applied level, we are now extending \vago\ with additional markers of subjectivity besides adjectives, in particular with explicit markers (first-person pronouns, exclamation marks), but also with more contextual features (direct vs. reported discourse). Ultimately, our goal is to update its interface to help users have a reliable grasp of levels of objectivity and subjectivity in the texts they read in the media.



\section*{Acknowledgements}
\noindent We thank four anonymous reviewers for helpful comments and several colleagues for feedback. This work was carried out with the support of the programs {\small HYBRINFOX (ANR-21-ASIA-0003), FRONTCOG (ANR-17-EURE-0017 program)}, and {\small PLEXUS (Marie Sk\l odowska-Curie Action, Horizon Europe Research and Innovation Programme, grant agreement n°101086295}).

\bibliographystyle{abbrv}
\bibliography{biblio}

\end{document}